\documentclass[10pt,twocolumn,letterpaper]{article}

\usepackage{cvpr}
\usepackage{times}
\usepackage{epsfig}
\usepackage{graphicx}
\usepackage{amsmath}
\usepackage{amssymb}
\usepackage{cases}
\usepackage{booktabs} 
\usepackage{subfigure}
\usepackage{multirow}
\usepackage{authblk}

\usepackage[breaklinks=true,bookmarks=false]{hyperref}

\cvprfinalcopy 
\ifcvprfinal\fi
\def\cvprPaperID{2856} 

\begin{document}

\title{On Exploring Undetermined Relationships for Visual Relationship Detection}

\author[1]{Yibing Zhan}
\author[1]{Jun Yu\thanks{Jun Yu is the corresponding author}}
\author[1]{Ting Yu}
\author[2]{Dacheng Tao}
\affil[1]{Key Laboratory of Complex Systems Modeling and Simulation,}
\affil[ ]{School of Computer Science and Technology, Hangzhou Dianzi University, China}
\affil[2]{UBTECH Sydney AI Centre, School of Computer Science, FEIT, University of Sydney,}
\affil[ ]{Darlington, NSW 2008, Australia}
\renewcommand\Authands{ and }
\affil[ ]{{\{zybjy,yujun\}@hdu.edu.cn, yuting@zufedfc.edu.cn, dacheng.tao@sydney.edu.au}}

\maketitle
\thispagestyle{empty}
\begin{abstract}
In visual relationship detection, human-notated relationships can be regarded as determinate relationships. However, there are still large amount of unlabeled data, such as object pairs with less significant relationships or even with no relationships. We refer to these unlabeled but potentially useful data as undetermined relationships. Although a vast body of literature exists, few methods exploit these undetermined relationships for visual relationship detection.

In this paper, we explore the beneficial effect of undetermined relationships on visual relationship detection. We propose a novel multi-modal feature based undetermined relationship learning network (MF-URLN) and achieve great improvements in relationship detection. In detail, our MF-URLN automatically generates undetermined relationships by comparing object pairs with human-notated data according to a designed criterion. Then, the MF-URLN extracts and fuses features of object pairs from three complementary modals: visual, spatial, and linguistic modals. Further, the MF-URLN proposes two correlated subnetworks: one subnetwork decides the determinate confidence, and the other predicts the relationships. We evaluate the MF-URLN on two datasets: the Visual Relationship Detection (VRD) and the Visual Genome (VG) datasets. The experimental results compared with state-of-the-art methods verify the significant improvements made by the undetermined relationships, e.g., the top-50 relation detection recall improves from 19.5\% to 23.9\% on the VRD dataset.
\end{abstract}

\section{Introduction}
\begin{figure}[!t]
\centering
{\includegraphics[width=0.97\linewidth]{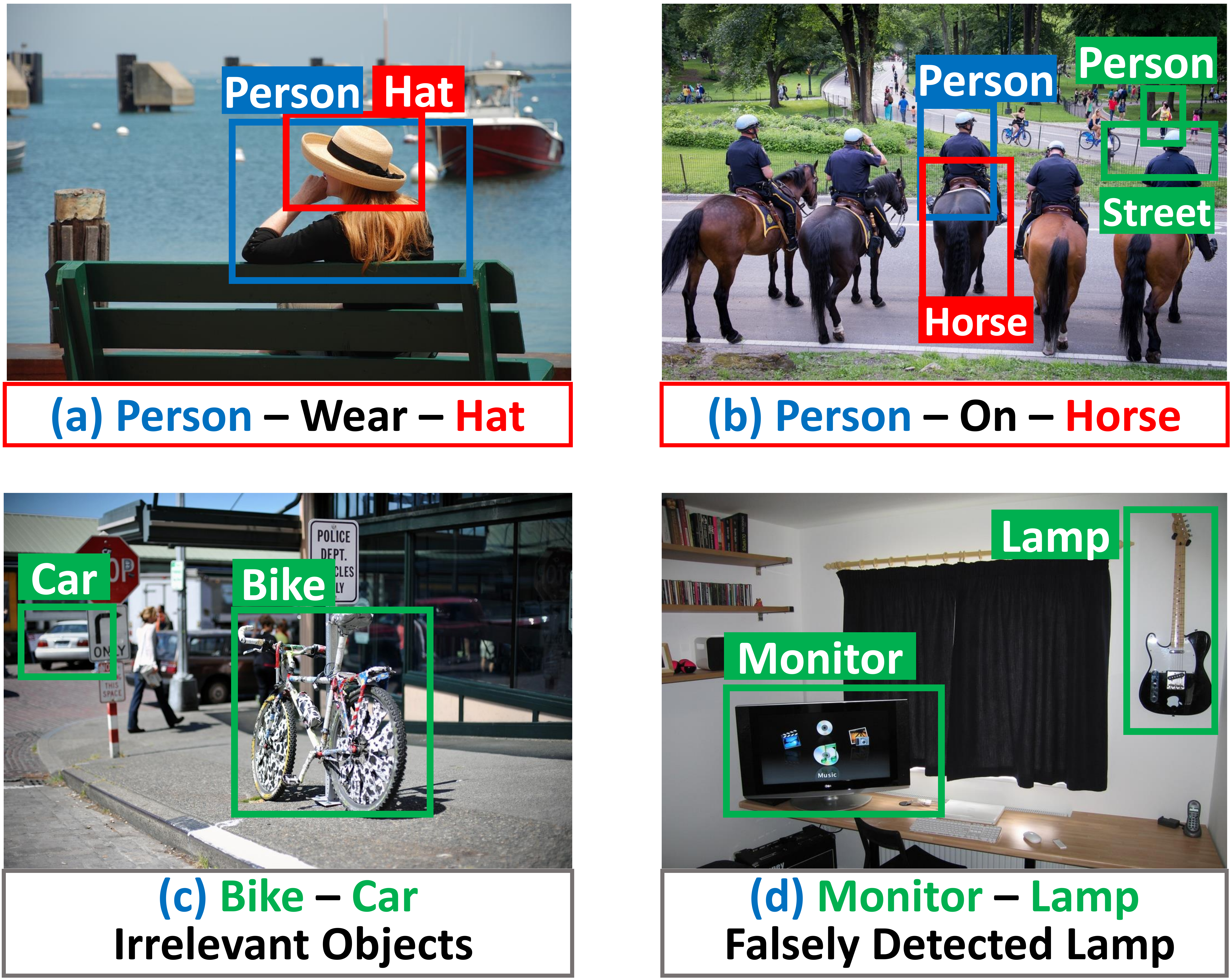}}
\caption{This figure shows four images with different relationships. (a) and (b) surround the descriptions of object pairs with determinate relationships. In addition, (b) presents an unlabeled relationship: person-on-street. (c) and (d) outline the descriptions of object pairs with undetermined relationships. In (d), the two objects are detected by Faster R-CNN \cite{ren2017faster}.}
\label{fig1}
\end{figure}
Visual relationships have been widely used in multiple image understanding tasks, such as object categorization \cite{galleguillos2008object}, object detection \cite{hu2018relation}, image segmentation \cite{gould2008multi}, image captioning \cite{fang2015from}, and human-object interactions \cite{gkioxari2017detecting}. Due to its wide applications, visual relationship detection has attracted increasingly more attention. The goal of visual relationship detection is to detect pairs of objects, meanwhile to predict the object pairs' relationships. In visual relationship detection, visual relationships are generally represented as subject-predicate-object triplets, such as person-wear-hat and person-on-horse, which are shown in Fig. \ref{fig1} (a) and (b). Since relation triplets are compositions of objects and predicates, their distribution is long-tailed. For $N$ objects and $M$ predicates, the number of all possible relation triplets is $O(N^2M)$. Therefore, taking relation triplets as a whole learning task requires a very large amount of labeling data \cite{desai2012detecting,sadeghi2011recognition}. A better strategy is to build separate modules for objects and predicates. Such a strategy reduces the complexity to $O(N+M)$ and increases the detection performance on large-scale datasets \cite{lu2016visual}. Even so, visual relationship detection is still data-hungry. Another solution is to obtain more human annotations. However, labeling relation triplets is truly expensive, as it requires tedious inspections of huge numbers of object interactions \cite{zhang2017ppr-fcn}. 

We notice that in addition to the human-notated relationships, there is still much unexploited data in images. We illuminate these data in Fig. \ref{fig1}. The human-notated relationships can be regarded as determinate relationships, such as Fig. \ref{fig1} (a) and (b). Contrarily, we refer to the other relationships constructed from unlabeled object pairs as undetermined relationships. These undetermined relationships include 1) object pairs with relationships but not labeled by humans, \eg, the unlabeled person-on-street shown in Fig. \ref{fig1} (b), and 2) object pairs with no relationships, such as Fig. \ref{fig1} (c). Further, object pairs with falsely detected objects are also classified as undetermined relationships, such as Fig. \ref{fig1} (d). Intuitively, these undetermined relationships can be used as complements of determinate relationships for the following reasons. First, they contain negative samples, such as object pairs without relationships and with falsely detected objects. Second, they reflect humans' dislike preferences, \eg , the less significant unlabeled relationships and relationships with unusual expressions (\eg, we prefer to say cup-on-table instead of table-under-cup), which are considered as undetermined relationships. Moreover, they require no human annotation and have beneficial regular effects for visual relationship detection \cite{zhang2017universum}.

Therefore, in this paper, we explore how to utilize these unlabeled undetermined relationships to improve relationship detection, and propose a multi-modal feature based undetermined relationship learning network (MF-URLN). In the MF-URLN, a generator is proposed to automatically produce useful undetermined relationships. Specifically, we use an object detector to detect objects, and two different objects compose an object pair; then, this object pair is compared with human-notated relationships using a designed criterion. Those object pairs without corresponding determinate relationships are classified as undetermined relationships. For each object pair, the MF-URLN extracts and fuses features from three different modals: the visual modal, the spatial modal, and the linguistic modal. These features comprehensively gather information on one relationship. Afterwards, the MF-URLN constructs two correlated subnetworks: one depicts the object pairs as either determinate or undetermined, and the other predicts the relationships. In addition, the second subnetwork uses information from the first subnetwork. The final relationships are decided according to the scores from the two subnetworks. We perform experiments on two relationship detection datasets, namely VRD \cite{lu2016visual} and VG \cite{krishna2017visual, zhang2017visual}, to verify the effectiveness of the MF-URLN. The experimental results demonstrate that the MF-URLN achieves great improvements on both datasets by using undetermined relationships, \eg, the top-50 phrase detection recall improves from 25.2\% to 31.5 \% in the VRD dataset.

Our contributions can be summarized as: 1) we explore undetermined relationships to improve visual relationship detection. We propose an automatic method to obtain effective undetermined relationships and a novel model to utilize these undetermined relationships for visual relationship detection. 2) We propose a novel and competitive visual relationship detection method, the MF-URLN, by using multi-modal features based on determinate and undetermined relationships. The experimental results, when compared with state-of-the-art methods, demonstrate the capability of the MF-URLN for visual relationship detection.

\section{Related Work}
{\bf Visual Relationship Detection.} Earlier works on visual relationship detection treated object and predicate detection as a single task. These methods required a large amount of training data, but could be applied only to limited situations \cite{desai2012detecting,sadeghi2011recognition}. Then, Lu \etal \cite{lu2016visual} proposed an efficient strategy for detecting the objects and predicates separately. Later, linguistic knowledge showed its power. Yu \etal \cite{yu2017visual} combined rich visual and linguistic representations using the teacher-student deep learning framework. Deep structural learning is another recent attempt. In \cite{liang2018visual}, a deep structural ranking model was proposed by integrating multiple cues to predict the relationships. The methods mentioned above contained two steps for objects and predicates. In contrast, other methods had end-to-end models. In \cite{zhang2017visual}, the authors proposed a end-to-end visual translation embedding network for relationship detection. Although the previous methods performed satisfactorily, few of them took undetermined relationships into account.
\begin{figure*}
\centerline{\includegraphics[width=0.95\linewidth]{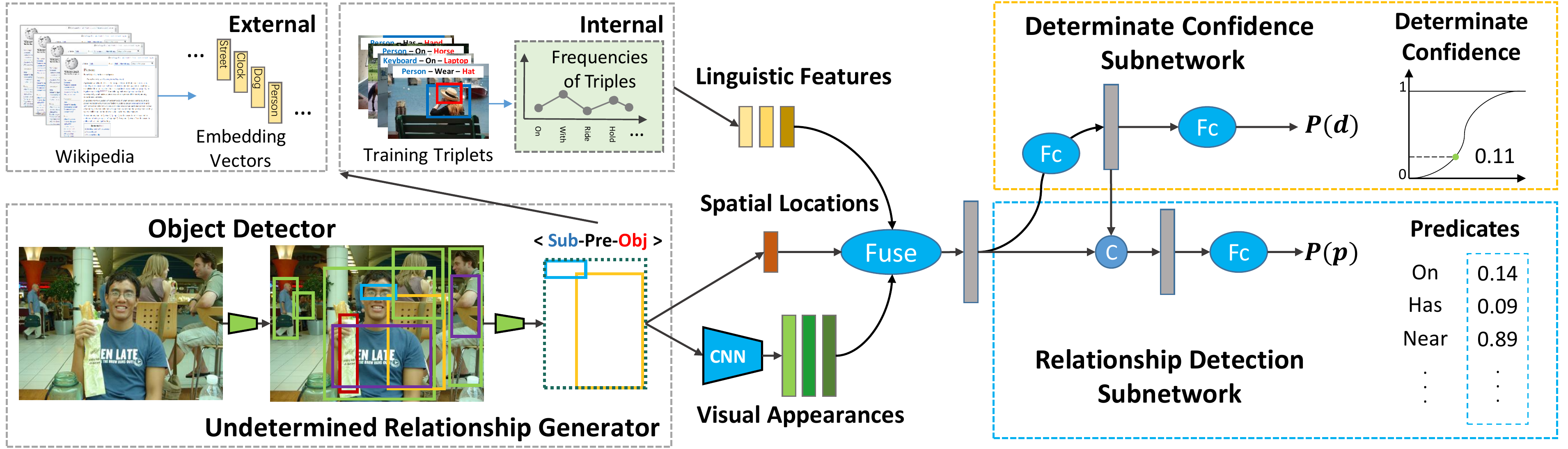}}
\caption{The framework of the MF-URLN. The MF-URLN detects objects through an object detector. Then, determinate and undetermined relationships are generated from the proposed generator. Afterwards, the MF-URLN extracts and fuses features from three modals to describe each object pairs. Finally, the relationships are predicted based on the determinate confidence subnetwork and the relationship detection subnetwork.}
\label{fig2}
\end{figure*}

{\bf Positive Unlabeled Learning.}
Utilization of undetermined relationships is related to positive unlabeled (PU) learning. PU learning refers to the task of learning a binary classifier from only positive and unlabeled data \cite{du2015convex}. PU learning has been used in a variety of tasks, such as matrix completion \cite{hsieh2015pu}, multi-view learning \cite{zhou2012multi}, and data mining \cite{li2009positive}. Most PU learning methods emphasize only binary classification \cite{platanios2017estimating}; \eg, \cite{sansone2018efficient} proposed an unlabeled data in sequential minimal optimization (USMO) algorithm to learn a binary classifier from an unlabeled dataset. However, visual relationship detection is a multi-label classification task. Therefore, this paper is one of the works for PU learning on multi-label tasks, similar to \cite{kaji2018multi,kanehira2016multi,zhang2017universum}. Following \cite{zhang2017universum}, we consider the beneficial effect of unlabeled relationships to improve visual relationship detection.
\section{MF-URLN}
Our novel multi-modal feature based undetermined relationship learning network (MF-URLN) is presented in this section. We suppose $s$, $p$, $o$, $d$, and $R$ to represent the subject, predicate, object, determinate confidence, and relationship, respectively. Therefore, the probabilistic model of the MF-URLN for visual relationship detection is defined as:
\begin{equation}\small
P(R) = P(p|s,o,d)P(d|s,o)P(s|B_{s})P(o|B_{o}).
\end{equation}

Here, $B_s$ and $B_o$ are two individual boxes for subject and object, which compose an object pair. $P(s|B_s)$ and $P(o|B_o)$ represent the subject box's and object box's probabilities for belonging to an object category. Since object detection is not a key topic in this paper, the MF-URLN directly obtains these object probabilities from an object detector, following \cite{lu2016visual,yu2017visual}. $P(d|s,o)$ represents the probability of the object pair having a determinate relationship; in other words, $P(d|s,o)$ reflects the probability of one object pair being manually selected and labeled. $P(p|s,o,d)$ is the probability of the object pair belonging to a predicate category. Note that only $P(s|B_s)$ and $P(o|B_o)$ are independent. The rest of the factors are correlated. $P(d|s,o)$ depends on the subject and the object, and $P(p|s,o,d)$ relies on the subject, the object, and the determinate confidence.

As shown in Fig. \ref{fig2}, the MF-URLN uses an object detector to detect objects and to provide the scores of $P(s|B_s)$ and $P(o|B_o)$. Then, an undetermined relationship generator is utilized. For training, object pairs are classified as determinate relationships and undetermined relationships. For testing, all object pairs are directly used as testing data. Finally, an undetermined relationship learning network is proposed to extract and fuse multi-modal features and to calculate the scores of $P(d|s,o)$ and $P(p|s,o,d)$. More details about the object detector, the undetermined relationship generator, and the undetermined relationship learning network are explained in the following subsections.
\subsection{Object Detector}
In the MF-URLN, we use the Faster R-CNN \cite{ren2017faster} with the VGG-16 network to locate and detect objects. Specifically, we first sample 300 proposed regions generated by the RPN with IoU $>$ 0.7. Then, after classification, we perform the NMS with IoU $>$ 0.4 on the 300 proposals. The retained proposals with category probabilities higher than 0.05 are regarded as the detected objects in one image. The Faster R-CNN with the VGG-16 is selected, because it is commonly used in visual relationship detection \cite{yang2018shuffle, yin2018zoom,zhang2017visual}. Note that the MF-URLN uses the same parameter settings of Faster R-CNN for different datasets, but these parameters can be adjusted to obtain better object proposals, following \cite{zhang2017visual,zhu2018deep}. In addition, the MF-URLN can be married to any object detector such as the fast RCNN \cite{girshick2014rich} and YOLO \cite{redmon2016you}.
\subsection{Undetermined Relationship Generator}
Different datasets of undetermined Relationships seriously affect the detection performance. Therefore, in this subsection, we introduce a fast manner to automatically generate useful undetermined relationships. 

Specifically, two different detected objects compose one object pair. Afterwards, all of the object pairs are compared with the manually annotated relationships (\emph{i.e.}, the ground truth) for categorization purposes. We suppose $l_s$, $l_p$, and $l_o$ represent the labels for the $s$, $p$, and $o$. $A$ represents the set of manually annotated relationships, and $D$ denotes the set of object pairs constructed from the detected objects. An object pair $(s_i, o_i)\in D$ is determinate only if $\exists (s_k, p_k, o_k)\in A$, in which $l_{s_i}=l_{s_k}$, $l_{o_i}=l_{o_k}$, $IoU(s_i, s_k) > 0.5$, and $IoU(o_i, o_k)>0.5$. In this way, $l_{p_i}=l_{p_k}$. Otherwise, $(s_i, o_i)$ is classified as an undetermined relationship, and $l_{p_i}$ is unknown and probably does not belong to any predicates. Here, $IoU(a, b)$ is the intersection over union (IoU) between the objects $a$ and $b$. 

We choose this generator because most of the generated undetermined relationships belong to the situations mentioned in the introduction. In addition, such a method also yields a good empirical performance. Note that we use the same object detector to detect objects and generate undetermined relationships, so as to make the generated undetermined relationships highly correlated with the detected objects of the MF-URLN. 
\subsection{Undetermined Relationship Learning Network}
The undetermined relationship learning network of the MF-URLN includes two parts: the multi-modal feature extraction network and the relationship learning network.
\subsubsection{Multi-modal Feature Extraction Network}
The MF-URLN  depicts a full picture of one relationship from features of three different modals: visual modal, spatial modal, and linguistic modal.

{\bf Visual Modal Features.} Visual modal features are useful to collect the category characteristics as well as the diversity of objects from the same category in different situations. Following \cite{yu2017visual}, the MF-URLN directly uses the VGG-16 with ROI pooling from the Faster R-CNN to extract the visual features from the individual boxes of subjects and objects, and the union box of subject and object in object pairs. In this way, our learning network is highly correlated with the object detector and the generated undetermined relationships. 

{\bf Spatial Modal Features.} Spatial modal features are complementary to visual modal features, because ROI pooling deletes the spatial information of object pairs. We suppose $(x^s_{min}, y^s_{min}, x^s_{max}, y^s_{max})$, $(x^o_{min}, y^o_{min}, x^o_{max}, y^o_{max})$, and $(x^u_{min}, y^u_{min}, x^u_{max}, y^u_{max})$ denote the locations of the subject box, the object box, and the union box of subject and object in one image, respectively. The spatial modal features are calculated as:
\begin{equation}\small
\begin{split}
[&\frac{x^s_{min}-x^u_{min}}{x^u_{max}-x^u_{min}}, \frac{y^s_{min}-y^u_{min}}{y^u_{max}-y^u_{min}},
\frac{x^s_{max}-x^u_{max}}{x^u_{max}-x^u_{min}}, \frac{y^s_{max}-y^u_{max}}{y^u_{max}-y^u_{min}}, \\
&\frac{x^o_{min}-x^u_{min}}{x^u_{max}-x^u_{min}}, \frac{y^o_{min}-y^u_{min}}{y^u_{max}-y^u_{min}},
\frac{x^o_{max}-x^u_{max}}{x^u_{max}-x^u_{min}}, \frac{y^o_{max}-y^u_{max}}{y^u_{max}-y^u_{min}} ].
\end{split}
\end{equation}

{\bf Linguistic Modal Features.} Linguistic modal features provide similarities among objects from linguistic knowledge, which are difficult to obtain from visual appearances and spatial locations. In the MF-URLN, object categories are obtained from the object detector; then, two kinds of linguistic modal features are extracted based on the labels associated with the classifier: external linguistic features and internal linguistic features. For external linguistic features, we employ the pretrained word2vec model of Wikipedia 2014 \cite{pennington2014glove} to extract the semantic representations of subject and object. However, such external linguistic features may contain noise because the training texts are not limited to relationships. Thus, the internal linguistic features are proposed as complements. For internal linguistic features, we count the frequencies of all relation triplets in the training set, and transform these frequencies into probability distributions according to the subject's and object's categories, based on Naive Bayes with Laplace smoothing  \cite{ng2002discriminative}. The Laplace smoothing is used to consider the zero-shot data influence \cite{lu2016visual}.

{\bf Feature Fusion.} In previous methods, the commonly used feature fusion method is directly concatenating features \cite{yu2017visual, zhang2017visual}. However, as different features' dimensions vary widely, high-dimensional features, such as the 4096-dimensional visual features, easily overwhelm low-dimensional features, such as the 8-dimensional spatial features. To alleviate such problems, in the MF-URLN, all individual features of the same modalities are transformed into the same dimensions and concatenated; then, these concatenated features of single modalities are again transformed into the same dimensions before being concatenated for multi-modal feature fusion.

\subsubsection{Relationship Learning Network}
Previous methods considered only determinate relationships. Contrarily, the MF-URLN predicts relationships from two kinds of data: determinate relationships and undetermined relationships. In detail, two subnetworks are proposed: a determinate confidence subnetwork and a relationship detection subnetwork.

{\bf Determinate Confidence Subnetwork.} The determinate confidence subnetwork decides the determinate confidence of an object pair, which reflects the probability of the object pair being manually selected and labeled. As shown in Fig. \ref{fig2}, our determinate confidence subnetwork uses multi-modal features. In the MF-URLN, we use sigmoid cross entropy loss \cite{de2005tutorial}, which is defined as:
\begin{equation}\label{eq0}\small
\text{ CE}(p,y)=
\begin{cases}
-\text{log}(p) & \text{if} \quad y =1 \\
-\text{log}(1-p) &\text{otherwise}.
\end{cases}
\end{equation}

We define $(s^d, p^d, o^d)$ as determinate relationships, and $(s^i, p^i, o^i)$ as undetermined relationships. For a determinate relationship, $y = 1$, and its determinate confidence loss is defined as:
$
L_{det}^d = \text{ CE}(P(d^{d}|s^{d},o^{d}),1).
$
For an undetermined relationship, $y = 0$, and its determinate confidence loss is defined as:
$
L_{det}^i = \text{ CE}(P(d^{i}|s^{i},o^{i}),0).
$
The final determinate confidence loss is weighted calculated as:
\begin{equation}\label{eq1}\small
L_{det} = L_{det}^d + \alpha L_{det}^i, 
\end{equation}
where $\alpha$ is the parameter to adjust the relative importance of undetermined relationships and determinate relationships. We believe that determinate relationships and undetermined relationships contribute equally to the determinate confidence loss and thus set $\alpha = 1$.

{\bf Relationship Detection Subnetwork.}
The relationship detection subnetwork predicts the relationship of all object pairs. As shown in Fig. \ref{fig2}, our relationship detection subnetwork depends on multi-modal features and the determinate confidence subnetwork. In this manner, the two subnetworks are correlated. In addition, the determinate confidence experimentally improves the relationship detection. 

Determinate relationships contain clear human-notated predicates. Therefore, the relationship detection loss from a determinate relationship is defined as:
\begin{equation}\small
L_{rel}^d =  \sum_{k=1}^M\text{CE}(P(p_k^d|s^d,o^d,d^d), y_k),
\end{equation}
where $p_k$ and $y_k$ are the $k$th predicate and the corresponding label. $y_k=1$ means the $k$th predicate's label is human-notated; otherwise, $y_k=0$. $M$ is the number of predicate categories.

The undetermined relationships with unlabeled predicates should have at least one predicate, whereas the undetermined relationships without any relationships or with falsely detected objects should have no predicates. There is currently no reliable methods to automatically label these undetermined relationships. Therefore, we treat these data as having no predicates, following \cite{zhang2017universum}. This method is naive but experimentally useful. The relationship detection loss from an undetermined relationship is defined as:
\begin{equation}\small
L_{rel}^i = \sum_{k=1}^M\text{CE}(P(p_k^i|s^i,o^i,d^i), 0).
\end{equation}

The relationship detection loss is finally calculated as:
\begin{equation}\label{eq2}\small
L_{rel} = L_{rel}^d + \lambda_1 L_{rel}^i. 
\end{equation}

Here, $\lambda_1$ is the parameter to adjust the relative significance of undetermined relationships and determinate relationships for the relationship detection loss.

{\bf Joint loss function.} 
Finally, a joint loss function is proposed to simultaneously calculate the determinate confidence loss and the relationship detection loss. The joint loss function is defined as follows:
\begin{equation}\label{eq3}\small
L = L_{rel} + \lambda_2 L_{det} .
\end{equation} 

Here, $\lambda_2$ is the parameter used to trade off between two groups of objectives: the determinate confidence loss and the relationship detection loss. By combining Eq. (\ref{eq1}), Eq. (\ref{eq2}), and Eq. (\ref{eq3}), the joint loss function is rewritten as:
\begin{equation}\small
L = L_{rel}^d +\lambda_1 L_{rel}^i + \lambda_2 L_{det}^d + \lambda_2 L_{det}^i. 
\end{equation}

\section{Experiments}
In this section, we conduct experiments to validate the effectiveness of the MF-URLN and the usefulness of undetermined relationships by answering the following questions. {\bf Q1:} is the proposed MF-URLN competitive, when compared with the state-of-the-art visual relation detection methods? {\bf Q2:} what are the influences of the features on the proposed MF-URLN? {\bf Q3:} are undetermined relationships beneficial to visual relationship detection? 

\subsection{Datasets, Evaluation Tasks, and Metrics}
{\bf Datasets.} Two public datasets are used for algorithm validation: the Visual Relationship Detection dataset \cite{lu2016visual} and the Visual Genome dataset \cite{krishna2017visual}.

The Visual Relationship Detection (VRD) dataset consists of 5,000 images with 100 object categories and 70 predicate categories. In total, the VRD contains 37,993 relationships with 6,672 types. The default dataset split includes 4,000 training images and 1,000 test images. There are 1,169 relation triplets that appear only in the test set, which are further used for zero-shot relationship detection. We split the default training images from the VRD into two parts: 3,700 images for training and 300 for validation.

The Visual Genome (VG) dataset is one of the largest relationship detection datasets. We note that there are multiple versions of VG datasets \cite{li2017vip-cnn,yin2018zoom,yu2017visual,zhang2017visual}. In this paper, we use the pruned version of the VG dataset provided by \cite{zhang2017visual}. This VG was also used in \cite{han2018visual,liang2018visual,yang2018shuffle,zhang2017ppr-fcn,zhu2018deep}. In summary, this VG contains 99,652 images with 200 object categories and 100 predicates. The VG contains 1,090,027 relation annotations with 19,561 types. The default dataset split includes 73,794 training and 25,858 testing images. We split the default training images from the VG into two parts: 68,794 images for training and 5,000 for validation.

{\bf Evaluation Tasks.} Three commonly used tasks are adopted: predicate detection, phrase detection, and relation detection, following \cite{yin2018zoom,yu2017visual}. In predicate detection, we are given an input image and ground truth bounding boxes with corresponding object categories. The outputs are predicates that describe each pair of objects. In phrase detection, we are given an input image. The output is a set of relation triplets and localization of the entire bounding box for each relation, which overlap at least 0.5 with the ground-truth joint subject and object boxes. In relation detection, we are given an input image. The output is a set of relation triplets and localization of individual bounding boxes for subjects and objects in each relation, which overlap at least 0.5 with the ground-truth subject and object boxes.

{\bf Evaluation Metrics.} We follow the precedent set by \cite{lu2016visual,yu2017visual} by using Recall as our evaluation metric. The top-$N$ recall is denoted as $R_{N}$. To be more specific, for one image, the outputs are the aggregation of the first $k$ top-confidence predicates from all the potential visual relation triplets in the image. The $R_{N}$ metric ranks all the outputs from an image and computes the recall of the top $N$. We use $R_{50}$ and $R_{100}$ for our evaluations. For both datasets, $k=1$. 

\subsection{Implementation Details}
We first train Faster R-CNN object detectors \cite{chen2017spatial} for the VRD and VG datasets individually. Then, undetermined and determinate relationships are generated from the proposed undetermined relationship generator. In our undetermined relationship learning network, the dimensions of all features' transforming layers are set as 500, following \cite{zhang2017visual}. The determinate confidence subnetwork includes two layers: a 100-dimensional fully connected feature-fusing layer and a sigmoid classification layer. The relationship detection subnetwork includes three layers: a concatenating layer for features of multi modals and the determinate confidence subnetwork, a 500-dimensional fully connected feature-fusing layer, and a sigmoid classification layer. We use the \begin{math}relu\end{math} function and the Adam Optimizer with an exponential decay learning rate to train the MF-URLN. For the VRD dataset, the initial learning rate is set at 0.0003, and it decays 0.5 every 4,000 steps. For the VG dataset, the initial learning rate is set at 0.0003, and it decays 0.7 every 35,000 steps. For the predicate detection task, we do not use undetermined relationships and set $\lambda_1$=$\lambda_2$=0. For the phrase and relation detection tasks, in each batch, the ratio of undetermined and determinate relationships is set at 3:1, following \cite{ren2017faster}. We set $\lambda_1$=0.5 and $\lambda_2$=1. The training sets are used to train the Faster R-CNN and the MF-URLN \footnote{Code available on https://github.com/Atmegal/}. The validation sets are used only to determine parameters.

\begin{table}[t]
  \caption{Performance comparison of visual relationship detection methods on the VRD dataset. Pre., Phr., and Rel. represent predication detection, phrase detection, and relation detection, respectively. ``-'' denotes that the result is unavailable.}\small
\centering
 \label{table1}{
\begin{tabular*}{\hsize}{@{}@{\extracolsep{\fill}}lccccc@{}}
    \toprule
  & Pre. &\multicolumn{2}{c}{Phr.}&\multicolumn{2}{c}{Rel.}\\
    	 & 	$R_{50/100}$&	$R_{50}$&	 $R_{100}$&	$R_{50}$&	 $R_{100}$\\
	\midrule 
    VRD-Full \cite{lu2016visual}&	47.9&16.2&	17.0	&13.9&	14.7	\\
	VTransE \cite{zhang2017visual}&	44.8		&19.4&	22.4	&	14.1&15.2	\\ 
	VIP-CNN \cite{li2017vip-cnn}&-			&22.8&	27.9&	17.3&	20.0	\\
	Weak-S \cite{peyre2017weakly-supervised}&	52.6		&17.9&	19.5&	15.8&	17.1\\	
	PPRFCN \cite{zhang2017ppr-fcn}&	47.4		&19.6&	23.2	&	14.4&	15.7	\\
	LKD:S \cite{yu2017visual}&		47.5		&19.2&	20.0&	16.6&	17.7\\
	LKD:T \cite{yu2017visual}&		54.1		&22.5&	23.6&	18.6&	20.6\\
	LKD:S+T \cite{yu2017visual}&	55.2		&23.1&	24.0&	19.2&	21.3\\
	DVSRL	 \cite{liang2017deep}&		-		&21.4&	22.6&	18.2&	20.8\\
	TFR \cite{hwang2018tensorize}&		52.3		&17.4&	19.1&	15.2&	16.8	\\
	DSL \cite{zhu2018deep}&		-		&22.7&	24.0&	17.4&	18.3 \\
	STA \cite{yang2018shuffle}&		48.0		&-	&		-&		-&		-	\\
	Zoom-Net \cite{yin2018zoom}&	50.7		&24.8&	28.1&	18.9	&	21.4\\
	CAI+SCA-M \cite{yin2018zoom}&	56.0	 	&25.2&	28.9	&	19.5&	22.4\\
	VSA \cite{han2018visual}&		49.2			&19.1	&21.7	&16.0&	17.7\\
	\midrule
	MF-URLN&	\textbf{58.2}&	\textbf{31.5}&	\textbf{36.1}&\textbf{23.9}&	\textbf{26.8}\\
    \bottomrule
  \end{tabular*}}
\end{table}
\begin{table}[t]
  \caption{Performance comparison of six methods on the VG dataset. ``-'' denotes that the result is unavailable.}\small 
\centering
 \label{table2}{
\begin{tabular*}{\hsize}{@{}@{\extracolsep{\fill}}lcccccc@{}}
    \toprule
     &\multicolumn{2}{c}{Pre.} &\multicolumn{2}{c}{Phr.}&\multicolumn{2}{c}{Rel.}\\
    		& 	$R_{50}$&	 $R_{100}$&	$R_{50}$&	 $R_{100}$&$R_{50}$&	 $R_{100}$\\
	\midrule 
VTransE \cite{zhang2017visual}&	62.6&	62.9&	9.5&	10.5&	5.5&	6.0\\
PPRFCN \cite{zhang2017ppr-fcn}&	64.2&	64.9&	10.6	&11.1&	6.0&	6.9\\
DSL \cite{zhu2018deep}&	-&	-&	13.1&	15.6&6.8&	8.0\\
STA \cite{yang2018shuffle}	&	62.7&	62.9&	-&	-&-&	-\\
VSA \cite{han2018visual}&	64.4&	64.5&	9.7&	10.0&6.0&	6.3\\
	\midrule
MF-URLN&	\textbf{71.9}&	\textbf{72.2}	&	\textbf{26.6}&	\textbf{32.1}&\textbf{14.4}&	\textbf{16.5}\\

    \bottomrule
  \end{tabular*}}
\end{table}
\subsection{Performance Comparisons (Q1)}
In this subsection, we compare the MF-URLN with state-of-the-art relationship detection models to show the competitiveness of the MF-URLN. We first compare the proposed MF-URLN with fifteen methods on the VRD dataset. The fifteen methods include: linguistic knowledge methods, such as VRD-Full \cite{lu2016visual}, LKD: S  \cite{yu2017visual}, LKD: T  \cite{yu2017visual}, and LKD: S+T  \cite{yu2017visual};  end-to-end network methods, such as VTransE \cite{zhang2017visual}, VIP-CNN \cite{li2017vip-cnn}, DVSRL \cite{liang2017deep}, and TFR \cite{hwang2018tensorize}; deep structural learning methods, such as DSL \cite{zhu2018deep}; and some other visual relationship detection methods, such as Weak-S \cite{peyre2017weakly-supervised}, PPRFCN \cite{zhang2017ppr-fcn}, STA \cite{yang2018shuffle}, Zoom-Net\cite{yin2018zoom}, CAI+SCA-M \cite{yin2018zoom}, and VSA \cite{han2018visual}. These methods encompass distinct and different properties. The results are provided in Table \ref{table1}\footnote{In predicate detection, $R_{50}$=$R_{100}$, because there are not enough objects in ground truth to produce more than 50 pairs.\label{footnote1}}. The best methods are highlighted in bold font. From Table \ref{table1}, it is apparent that the MF-URLN outperforms all the other methods in all tasks. In predicate detection, the MF-URLN outperforms the second-best competitor by 3.9\% on $R_{50/100}$. In phrase detection, compared with the second-best method, the MF-URLN increases $R_{50}$ and $R_{100}$ by 25.0\% and 24.9\%, respectively. In relation detection, compared with the second-best method, the MF-URLN improves upon $R_{50}$ and $R_{100}$ by 22.6\% and 19.6\%, respectively. These high performances demonstrate the capacity of the MF-URLN for relationship detection. 

Table \ref{table2} provides the performance of the MF-URLN and five competitive methods (VtransE, PPRFCN, DSL, STA, and VSA-Net) on the VG dataset. The results from the other methods are not provided, because these methods were not tested on the version of VG dataset \cite{zhang2017visual} in their corresponding papers. In Table \ref{table2}, the top methods are highlighted in boldface. The MF-URLN performs the best in all of the tasks no matter the evaluation criteria. For predicate detection, The MF-URLN yields 11.6\% and 11.2\% gains on $R_{50}$ and $R_{100}$ for predicate detection, 103.1\% and 105.8\% on $R_{50}$ and $R_{100}$ for phrase detection, and 111.8\% and 106.3\% on $R_{50}$ and $R_{100}$ for relation detection, respectively. These improvements verify that the MF-URLN can be applied to large-scale datasets with complex situations.

We also provide the zero-shot detection performances of ten methods in Table \ref{table3}\textsuperscript{\ref{footnote1}} to evaluate the ability of the MF-URLN to tackle zero-shot data. These methods include: VRD-Full, VtransE, Weak-S, LKD: S, LKD: T, DVSRL, TFR, the MF-URLN, and the MF-URLN-IM. 
The remaining methods are not compared, because the corresponding papers did not consider zero-shot relationship detection. Here, the MF-URLN-IM is the MF-URLN with the inferring model\footnote{The inferring model is explained in the supplement materials.}. In Table \ref{table3}, the best methods are highlighted in boldface. The MF-URLN still performs nearly the best on predicate detection. However, on phrase detection and relation detection, the MF-URLN does not perform as well. This result may occur because some unseen determinate relationships have been mistakenly classified as undetermined and consequently influence the zero-shot detection performance. The MF-URLN-IM improves the performance of the MF-URLN, because of the inferring model. However, this inferring model is not proper for seen data. The $R_{50/100}$ predicate relation on the VRD of the MF-URLN-IM is only 57.2. Better strategies to generate and utilize undetermined relationships are still necessary.

\begin{table}[t]
  \caption{Performance comparison on the zero-shot set of the VRD dataset. ``-'' denotes that the result is unavailable.}\small
\centering
 \label{table3}{
\begin{tabular*}{\hsize}{@{}@{\extracolsep{\fill}}lccccc@{}}
    \toprule
     &Pre. &\multicolumn{2}{c}{Phr.}&\multicolumn{2}{c}{Rel.}\\
    		& 	$R_{50/100}$&	$R_{50}$&	 $R_{100}$&	$R_{50}$&	 $R_{100}$\\
	\midrule 
VRD-Full \cite{lu2016visual}& 	12.3& 5.1& 	5.7& 4.8& 5.4\\
VTransE \cite{zhang2017visual}& 	 	-& 2.7& 	3.5& 	1.7& 	2.1\\
Weak-S \cite{peyre2017weakly-supervised}& 	 21.6& 6.8& 	7.8& 6.4& 	7.4\\
LKD:S \cite{yu2017visual}	& 17.0&\textbf{10.4}& \textbf{10.9}&\textbf{8.9}& 	\textbf{9.1}\\
LKD:T \cite{yu2017visual}& 	 	8.8& 	6.5	& 6.7	& 6.1	& 6.4\\
DVSRL	 \cite{liang2017deep} & 	 	-& 9.2& 	10.3	&7.9& 	8.5\\
TFR \cite{hwang2018tensorize}	&  	17.3	& 5.8& 	7.1	& 5.3& 	6.5 \\
STA \cite{yang2018shuffle}	&  	20.6	&-& 	-	&-& 	-\\
	\midrule 
MF-URLN& 	26.9	& 5.9&	7.9& 4.3&	5.5	 \\
MF-URLN-IM& 	\textbf{27.2}	& 6.2&	9.2& 4.5&	6.4	 \\
    \bottomrule
  \end{tabular*}}
\end{table}
\subsection{Discussion of Multi-modal Features (Q2)}
In this subsection, the effects of multi-modal features on the MF-URLN are discussed. The MF-URLN is compared with its eight variants by conducting predicate and relation detection on the VRD dataset. These eight variants include three baselines of single-modal features, the ``$V$'', the ``$S$'', and the ``$L_{ex,in}$'', in which the MF-URLN uses only visual modal features, spatial modal features, and linguistic modal features, respectively; three methods of bi-modal features, the ``$V$+$S$'', the ``$V$+$L_{ex,in}$'', and the ``$L_{ex,in}$+$S$'', in which the  MF-URLN uses visual and spatial modal features, visual and linguistic modal features, and linguistic and spatial modal features, respectively; and two methods of multi-modal features, the ``$V$+$S$+$L_{in}$'' and the ``$V$+$S$+$L_{ex}$'', in which the MF-URLN uses internal and external linguistic features, respectively. Additionally, we discuss two kinds of feature fusion methods: the ``Transformation'', in which methods transform features into the same dimensions before concatenating, and the ``Concatenating'', in which methods directly concatenate features. Note that concatenating methods have the same layer number and layer dimensions as their corresponding transforming methods to ignore the improvements caused by deepening the net. 

The performances of all compared methods are provided in Table \ref{table4}. We draw the following conclusions. 1) By comparing the MF-URLN with methods having different features, we can see that the MF-URLN obtains the best performance. Features from different modalities are complementary and all contribute to the performance of the MF-URLN. 2) By comparing concatenating methods with transforming methods, we can conclude that directly concatenating features is a low effective feature fusion strategy and that transforming all features into the same dimensions improves the performance. However, we notice that concatenating methods slightly outperforms transforming methods on single modal features. Better feature fusion strategy is still necessary and remains a future topic.
\begin{table}[t]
  \caption{$R_{50}$ predicate detection and relation detection of the MF-URLN and its eight variants on the VRD dataset.}\small
\centering
  \label{table4}{
\begin{tabular*}{\hsize}{@{}@{\extracolsep{\fill}}lcccc@{}}
    \toprule
     &\multicolumn{2}{c}{Transforming} &\multicolumn{2}{c}{Concatenating}\\
	 & Pre.& Rel.& Pre.&Rel.\\
    \midrule 
 Baseline: $V$ &52.29&22.64	&53.01&22.85\\
 Baseline: $L_{ex,in}$&53.39&	18.49&53.94&18.07\\
 Baseline: $S$&43.43&	17.94&43.44&17.95\\
$V$+$S$&54.66&23.15	&52.36&22.75\\   
$V$+$L_{ex,in}$&57.27&23.21	&55.45&22.62\\  
$L_{ex,in}$+$S$&57.10&23.67	&\textbf{56.04}&\textbf{23.29}\\  
$V$+$S$+$L_{in}$ &56.87&23.15	&53.25&22.51\\ 
$V$+$S$+$L_{ex}$ &57.69&23.50	&55.29&22.83\\ 
MF-URLN&\textbf{58.22}&\textbf{23.89}	&55.77&22.61\\
    \bottomrule
  \end{tabular*}}
\end{table}
\subsection{Analysis of undetermined Relationships (Q3)}
In this subsection, we first validate the usefulness of undetermined relationships in visual relationship detection. We compare the MF-URLN with its three variants by conducting relation detection on the VRD dataset. The three variants include the baseline ``MFLN'', which is the MF-URLN without using undetermined relationships; the ``MFLN-Triplet NMS'', which is the MFLN with triplets NMS \cite{li2017vip-cnn}; and the ``MFLN-Pair Filtering'', which is the MFLN that uses pair filtering \cite{dai2017detecting}. The triplets NMS and pair filtering have both been proposed to delete negative object pairs. The performances of these four methods are compared in Table \ref{table5}. We see that ``MFLN-Triplet NMS'' decreases the performance of ``MFLN'', partly because the NMS has already been used in our object detector. ``MFLN-Pair Filtering'' increases the performance of ``MFLN'', because it excludes some undetermined object pairs. The MF-URLN achieves the best performance. The improvement of the MF-URLN verify the utility of undetermined relationships for visual relationship detection. 

\begin{table}[t]
  \caption{Relation detection of four methods on the VRD dataset.}\small
\centering
  \label{table5}{
\begin{tabular*}{\hsize}{@{}@{\extracolsep{\fill}}lcccc@{}}
    \toprule
	 & \multicolumn{2}{c}{Entire Set} &\multicolumn{2}{c}{ Unseen Set}\\

 & $R_{50}$ & $R_{100}$ &$R_{50}$ &$R_{100}$\\
  \midrule 
Baseline: MFLN& 17.36&	21.76	&4.02	&4.96\\
 MFLN-Triplet NMS&15.53&	17.95&3.76&	4.19\\
 MFLN-Pair Filtering&21.58&	23.39&3.59&3.93\\
MF-URLN&\textbf{23.89}&	\textbf{26.79}&	\textbf{4.28}&	\textbf{5.47}\\
    \bottomrule
  \end{tabular*}}
\end{table}
\begin{figure}[!t]
\centering
\subfigure[The effects of  $\lambda_1$ and $\lambda_2$.] {\includegraphics[width=0.47\linewidth]{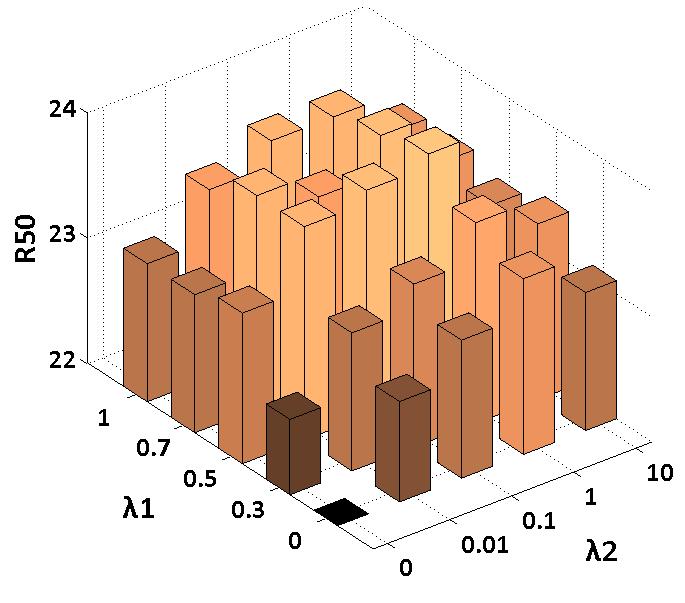}}
\hspace{0.03px}
\subfigure[MF-URLN vs MF-URLN-NS.] {\includegraphics[width=0.47\linewidth]{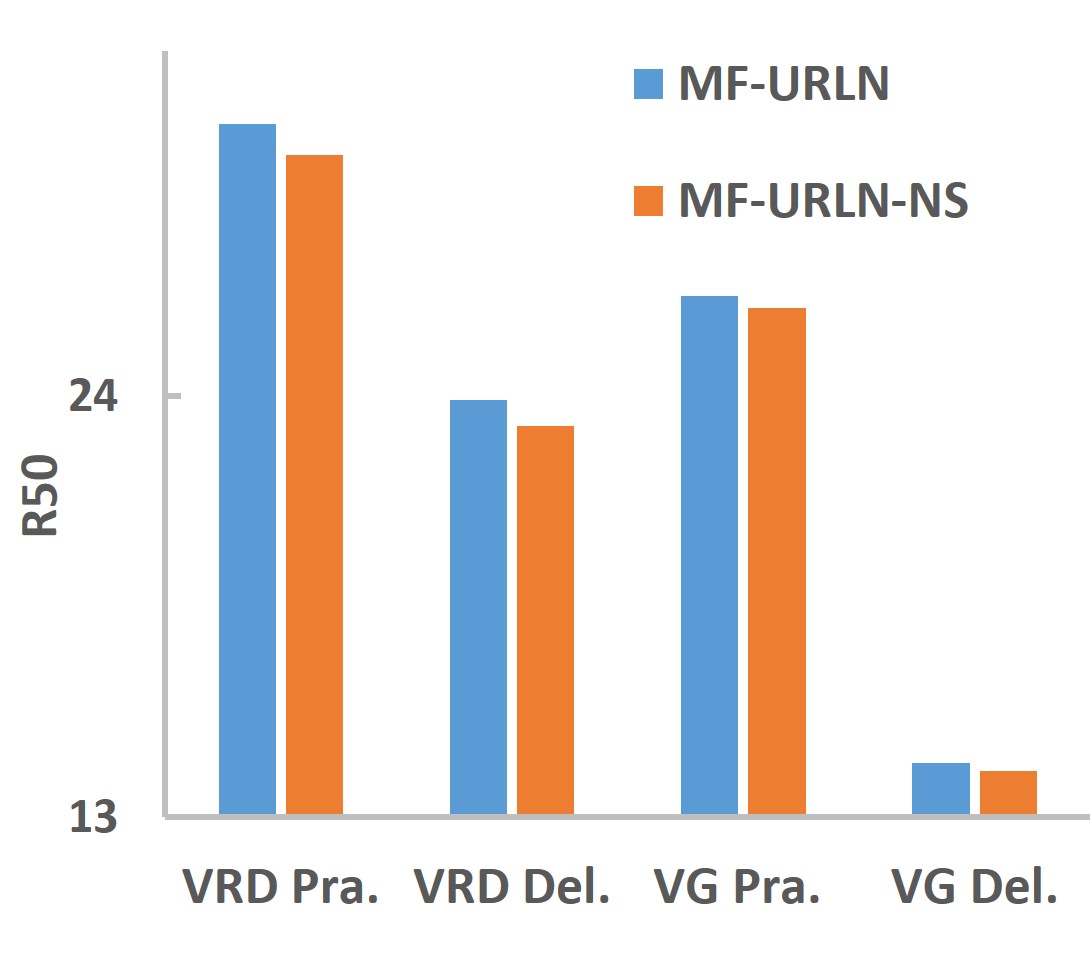}}
\caption{(a) Detection performance of the MF-URLN with different values of modal parameters: $\lambda_1$ and $\lambda_2$. (b) Performance comparison of the MF-URLN and the MF-URLN-NS.}
\label{fig3}
\end{figure}
Then, we discuss the beneficial effects of undetermined relationships on the MF-URLN by studying the performance impacts of the modal parameters, $\lambda_1$ and $\lambda_2$. The $R_{50}$ relation detection on the VRD dataset is used as the evaluation criterion. Specifically, the evaluation is conducted by changing one of the observed parameters while fixing the other, as in \cite{zhan2018comprehensive}. We set the ranges of $\lambda_1$ and $\lambda_2$ as \{0, 0.3, 0.5, 0.7, 1\} and \{0, 0.01, 0.1, 1, 10\}, respectively. Fig. \ref{fig3} (a)\footnote{$\lambda_1$=0 and $\lambda_2$=0 denote the MF-URLN without using undetermined relationships, the result is 17.36.} shows the performance. We can observe that when $\lambda_1$=0.5 and $\lambda_2$=1, the MF-URLN yields the best performance. $\lambda_1$=0.5 reveals that labeling undetermined relationships as having no predicates is a useful strategy; undetermined relationships do have beneficial regular effects on the relationship detection. $\lambda_2$=1 reveals that both of the subnetworks contribute to the detection of the MF-URLN.

Next, we show the benefit of sharing information from the determinate confidence subnetwork with the relationship detection subnetwork. We compare the MF-URLN with the MF-URLN-NS, in which the relationship detection subnetwork does not use information from the confidence subnetwork. The comparing performances are shown in Fig. \ref{fig3} (b). It can be seen that the MF-URLN performs better. This result verifies that the determinate confidence also contributes to the relationship detection. 

\begin{figure}[t]
\centerline{\includegraphics[width=0.92\linewidth]{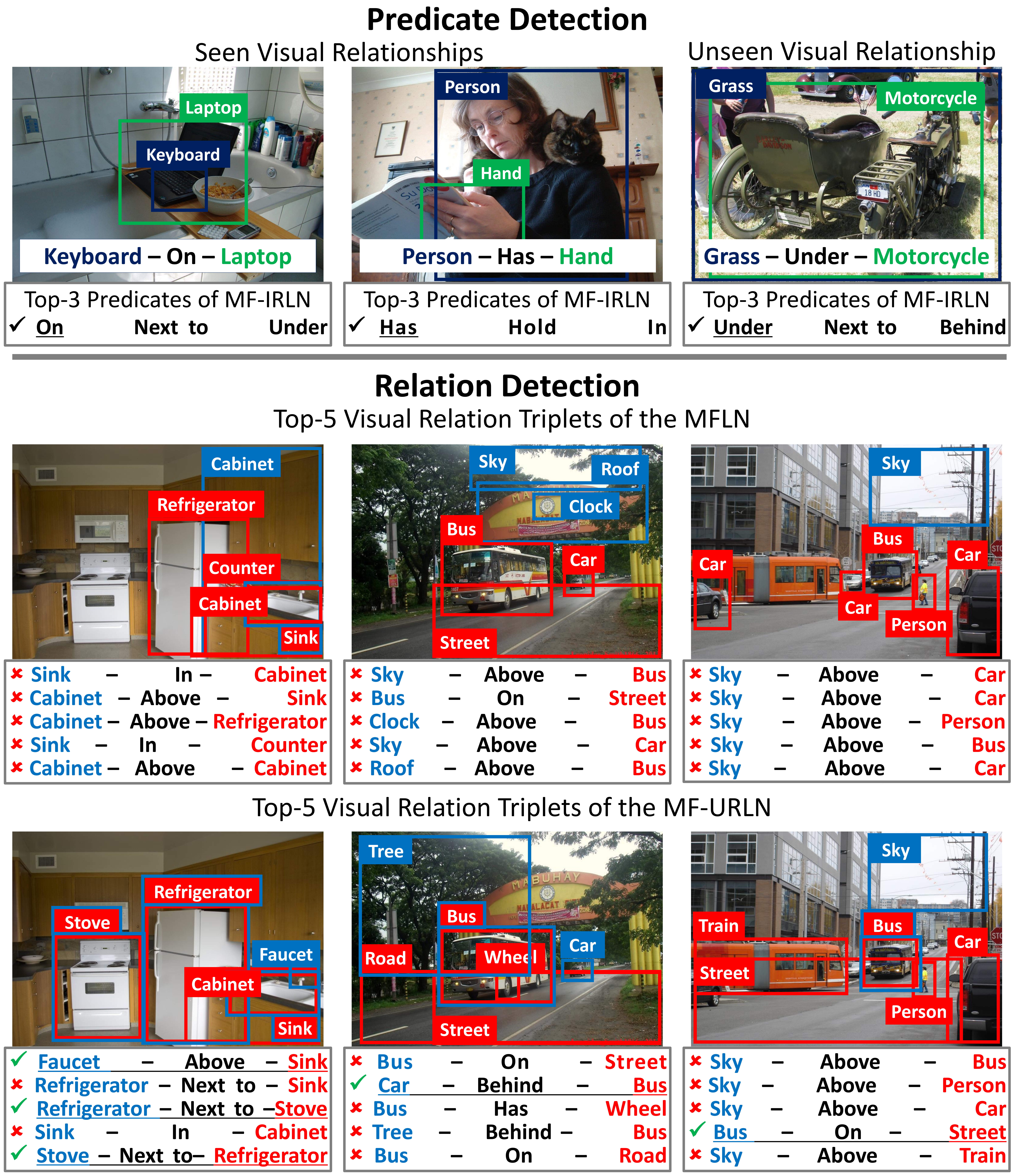}}
\caption{Visualization of detection results. For predicate detection, the top-3 predicates of the MF-URLN are provided. For relation detection, the top-5 triplets of the MFLN and the MF-URLN are provided. The $\surd$ represents the correct results. }
\label{fig4}
\end{figure}

We also provide the quantitative performances of the MF-URLN in Fig. \ref{fig4}. For predicate detection, it can be seen that the MF-URLN yields accurate predictions, which reveals the ability of the MF-URLN. For relation detection, the MF-URLN obtains much better detections than the MFLN. Utilization of undetermined relationships highlights object pairs with determinate relationships.
\section{Conclusion}
In this paper, we explore the role of undetermined relationships in visual relationship detection. And accordingly, we propose a novel relationship detection method, MF-URLN, which extracts and fuses multi-modal features based on determinate and undetermined relationships. The experimental results, when compared with state-of-the-art methods, demonstrate the competitiveness of the MF-URLN and the usefulness of undetermined relationships. Our future works include better utilizing undetermined relationships for relationship detection and promoting undetermined relationships to scene graph generation \cite{hwang2018tensorize, xu2017scene,zellers2018neural}.
\section{Supplement Materials}
\subsection{MF-URLN-IM}
To tackle the problem of zero-shot learning, we propose a multi-modal feature based undetermined relationship learning network with inferring model (MF-URLN-IM). The inferring model is inspired by humans' natural gift for inference wherein a person is able to predict the relationship between two objects from partial information obtained from learned object pairs. This process is illuminated in Fig. \ref{fig1}. Therefore, when encountering unseen relationships, MF-URLN-IM still performs robustly, according to the information obtained from the individual subjects and objects. 

Specifically, MF-URLN-IM has three separate types of relationship learning networks: a union relationship learning network, a subject relationship learning network, and a object relationship learning network. All relationship learning networks share the same architecture of MF-URLN, except they have different input features. The union relationship learning network includes all of the features. The subject relationship learning network includes the subjects' visual features, subjects' external linguistic features, and spatial features. The object relationship learning network includes the objects' visual features, objects' external linguistic features, and spatial features. The visual features of union boxes and internal linguistic features are not used in the subject and object relationship learning network because these two features contain both subject and object information. A joint loss function is used to simultaneously train the three relationship learning networks. The joint loss function is defined as:
\begin{equation}
L = L_{sub+obj}+L_{sub}+L_{obj},
\end{equation}where $ L_{sub+obj}$, $L_{sub}$, and $L_{obj}$ represents the loss functions for the union, subject, and object relationship learning network, respectively.

By using this joint loss function, the three relationship learning networks can share the same parameters as in previous modules. The final relationship is predicted by calculating the geometric average of the predictions from the three networks. This is calculated as:
\begin{equation}
P(R) = P(R|u)\cdot P(R|s) \cdot P(R|o).
\end{equation}
where $P(R|u)$, $P(R|s)$, and $P(R|o)$ represents the relationship probabilities of the union, subject, and object relationship learning network, respectively.

Table \ref{table1} compares performances of MF-URLN and MF-URLN-IM. As shown, MF-URLN-IM outperforms MF-URLN in all tasks. This results reveal the potential usefulness of the inferring model for visual relationship detection.
\begin{table}[t]
\caption{Performance comparison on the zero-shot set of the VRD dataset.}\small
\centering
\label{table1}{
\begin{tabular*}{\hsize}{@{}@{\extracolsep{\fill}}lccccc@{}}
\toprule
&Pre. &\multicolumn{2}{c}{Phr.}&\multicolumn{2}{c}{Rel.}\\
& $R_{50/100}$& $R_{50}$& $R_{100}$& $R_{50}$& $R_{100}$\\
\midrule 

MF-URLN& 26.9 & 5.9& 7.9& 4.3& 5.5 \\
MF-URLN-IM& \textbf{27.2} & \textbf{6.2}& \textbf{9.2}& \textbf{4.5}& \textbf{6.4} \\
\bottomrule
\end{tabular*}}
\end{table}

\subsection{More Discussion of Undetermined Relationships}
In this subsection, more qualitative results are provided.
\begin{figure}[t]
\centerline{\includegraphics[width=0.92\linewidth]{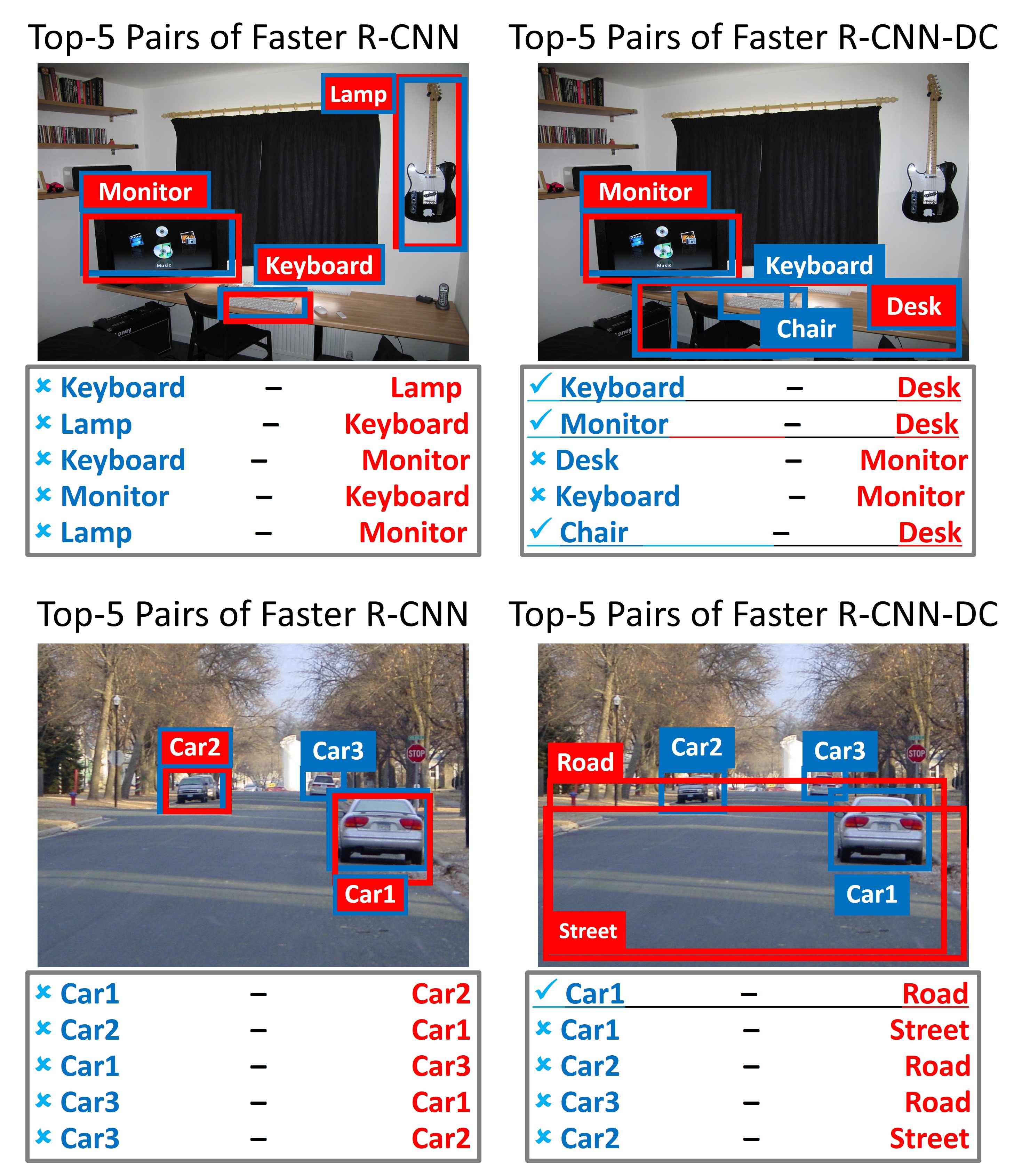}}
\caption{The top-5 detected object pairs of Faster R-CNN and Faster R-CNN with determinate confidence scores (Faster R-CNN-DC). In Faster R-CNN, object pairs are ranked by the product of subject boxes' and object boxes' probabilities. In Faster R-CNN-DC, object pairs are ranked by the product of subject boxes', object boxes', and determinate confidence probabilities. The $\surd$ represents the manual-labeled object pairs.}
\label{fig2}
\end{figure}
Fig. \ref{fig2} provides two examples of top-5 object pairs detected by Faster R-CNN and Faster R-CNN-DC. Faster R-CNN-DC refers to the method, which uses Faster R-CNN to detect objects and uses determinate confidence subnetwork to produce determinate confidence scores for object pairs. In Faster R-CNN, object pairs are ranked by the product of subject boxes' and object boxes' probabilities. In Faster R-CNN-DC, object pairs are ranked by the product of subject boxes', object boxes', and determinate confidence probabilities. As shown in Fig. \ref{fig2}, Faster R-CNN-DC outperforms Faster R-CNN in both examples. By adding determinate confidence subnetwrok, the object pairs with determinate relationships are highlighted. These highlighted determinate relationships results in better performance of visual relationship detection. Since determinate confidence subnetwork is trained based on undetermined relationships, the advantage of determinate confidence subnetwork again confirms the necessity and usefulness of undetermined relationships in visual relationship detection. In addition, in the upper example of Faster R-CNN of Fig. \ref{fig2}, the guitar is falsely detected as a lamp by the Faster R-CNN. Such mistake negatively influences the performance of a visual relationship detection method. Contrarily, in the example of Faster R-CNN-DC, we observe that the object pairs that contain falsely detected objects are ignored. This is because the object pairs with falsely detected objects are labeled as undetermined relationships. Using undetermined relationships in visual relationship detection alleviate the problem of falsely detected objects to some extent.
\begin{figure}[!t]
\centering
\subfigure[The failed case of predicate detection.] {\includegraphics[height=0.32\linewidth]{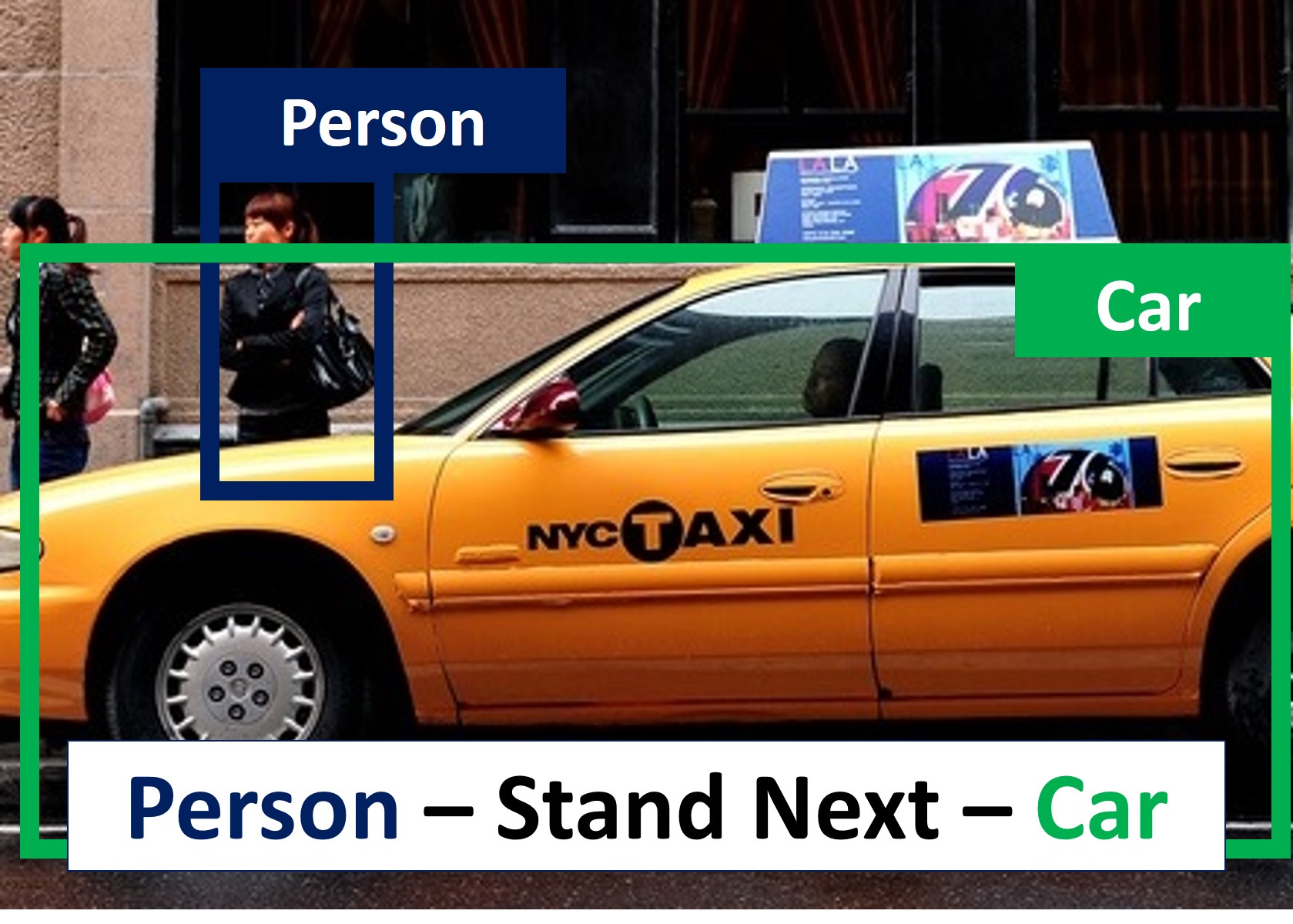}}
\hspace{0.05px}
\subfigure[The failed case of relation detection.] {\includegraphics[height=0.32\linewidth]{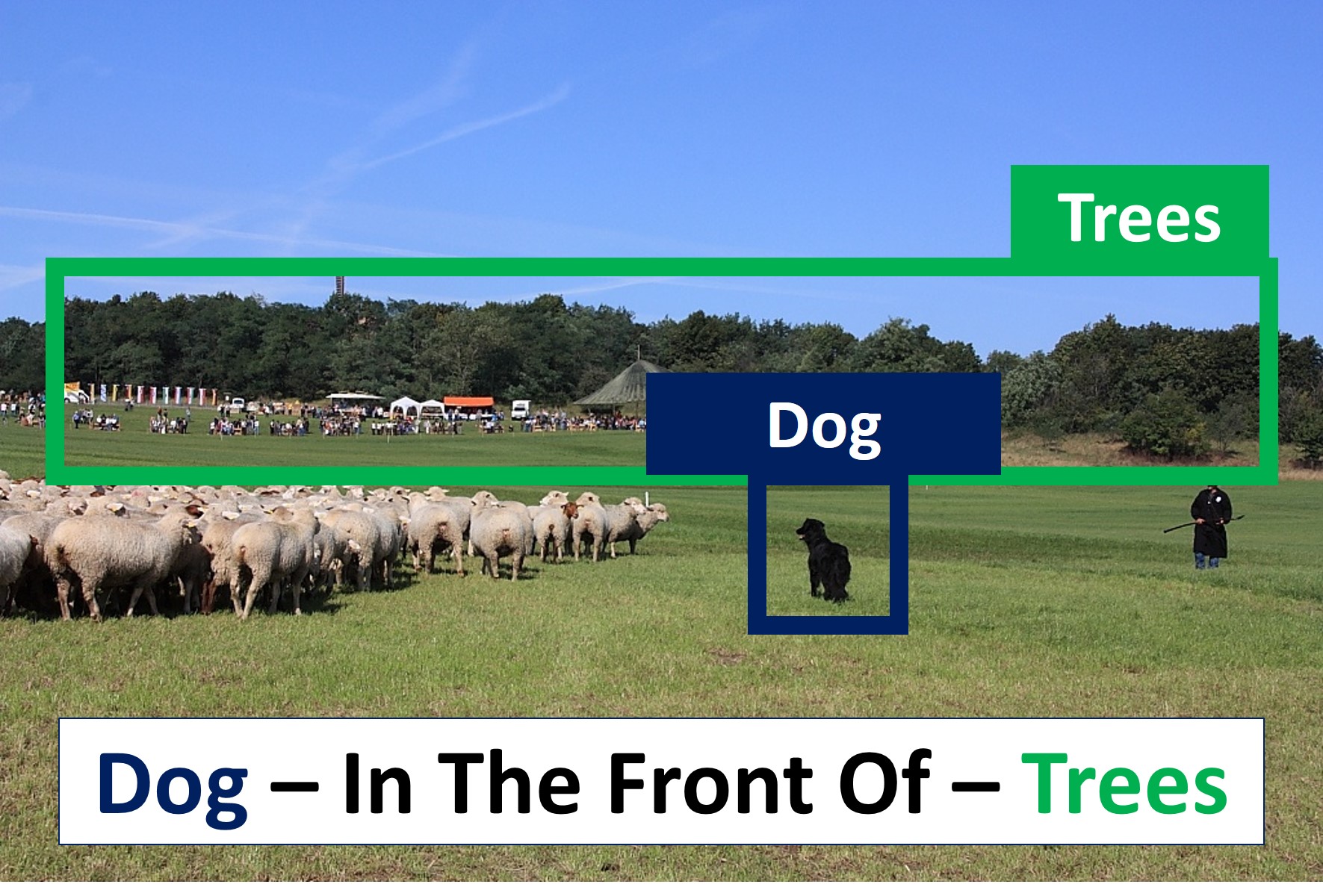}}
\caption{Two failed cases of MF-URLN. (a) The failed case of predicate detection. The predicates of both MF-URLN and MFLN are Person-Sit On-Car. (b) The failed case of relation detection. Both of MF-URLN and MFLN correctly predict the predicates. In MFLN, the relationship dog-in the front of-trees is the No.11 recall of relation detection. In MF-URLN, the relationship is No.233 recall of relation detection.}
\label{fig3}
\end{figure}

Fig. \ref{fig3} presents two failed cases of MF-URLN. Fig. \ref{fig3} (a) is a failed case of predicate detection. The detected predicate of both MF-URLN and MFLN for the given person and car is ``sit on''. This failure is caused because legs of the person are obscured by the car and it is difficult for MF-URLN and MFLN to identify the posture of the person. 
Fig. \ref{fig3} (b) provides a failed case of relation detection. Both of MF-URLN and MFLN predict correct predicates between the dog and the trees. In MFLN, the relationship dog-in the front of-trees is in the top-50 recall of relation detection, whereas in MF-URLN, the relationship is not in the top-50. This is because the relationship dog-in the front of-trees has low probability score of determinate confidence. The failure of relation detection indicate that better strategies to generate and utilize undetermined relationships are still necessary.

{\bf Acknowledgment.}
This work was supported in part by the National Natural Science Foundation of China under Grant No.: 61836002 and 61622205, and in part by the Australian Research Council Projects FL-170100117, DP-180103424, and IH-180100002.

{\small
\bibliographystyle{ieee}
\bibliography{egbib}

\begin{thebibliography}{10}\itemsep=-1pt

\bibitem{chen2017spatial}
X.~Chen and A.~Gupta.
\newblock Spatial memory for context reasoning in object detection.
\newblock {\em arXiv preprint arXiv:1704.04224}, 2017.

\bibitem{dai2017detecting}
B.~Dai, Y.~Zhang, and D.~Lin.
\newblock Detecting visual relationships with deep relational networks.
\newblock {\em Computer vision and pattern recognition}, pages 3298--3308,
  2017.

\bibitem{de2005tutorial}
P.-T. De~Boer, D.~P. Kroese, S.~Mannor, and R.~Y. Rubinstein.
\newblock A tutorial on the cross-entropy msethod.
\newblock {\em Annals of operations research}, 134(1):19--67, 2005.

\bibitem{desai2012detecting}
C.~Desai and D.~Ramanan.
\newblock Detecting actions, poses, and objects with relational phraselets.
\newblock In {\em European Conference on Computer Vision}, pages 158--172.
  Springer, 2012.

\bibitem{du2015convex}
M.~Du~Plessis, G.~Niu, and M.~Sugiyama.
\newblock Convex formulation for learning from positive and unlabeled data.
\newblock In {\em International Conference on Machine Learning}, pages
  1386--1394, 2015.

\bibitem{fang2015from}
H.~Fang, S.~Gupta, F.~Iandola, R.~K. Srivastava, L.~Deng, P.~Dollar, J.~Gao,
  X.~He, M.~Mitchell, J.~C. Platt, C.~L. Zitnick, and G.~Zweig.
\newblock From captions to visual concepts and back.
\newblock In {\em 2015 IEEE Conference on Computer Vision and Pattern
  Recognition (CVPR)}, volume~00, pages 1473--1482, June 2015.

\bibitem{sadeghi2011recognition}
A.~Farhadi and M.~A. Sadeghi.
\newblock Recognition using visual phrases.
\newblock In {\em CVPR 2011(CVPR)}, volume~00, pages 1745--1752, 06 2011.

\bibitem{galleguillos2008object}
C.~Galleguillos, A.~Rabinovich, and S.~Belongie.
\newblock Object categorization using co-occurrence, location and appearance.
\newblock In {\em Computer Vision and Pattern Recognition, 2008. CVPR 2008.
  IEEE Conference on}, pages 1--8. IEEE, 2008.

\bibitem{girshick2014rich}
R.~Girshick, J.~Donahue, T.~Darrell, and J.~Malik.
\newblock Rich feature hierarchies for accurate object detection and semantic
  segmentation.
\newblock In {\em Proceedings of the IEEE conference on computer vision and
  pattern recognition}, pages 580--587, 2014.

\bibitem{gkioxari2017detecting}
G.~Gkioxari, R.~Girshick, P.~Doll{\'a}r, and K.~He.
\newblock Detecting and recognizing human-object interactions.
\newblock {\em arXiv preprint arXiv:1704.07333}, 2017.

\bibitem{gould2008multi}
S.~Gould, J.~Rodgers, D.~Cohen, G.~Elidan, and D.~Koller.
\newblock Multi-class segmentation with relative location prior.
\newblock {\em International Journal of Computer Vision}, 80(3):300--316, 2008.

\bibitem{han2018visual}
C.~Han, F.~Shen, L.~Liu, Y.~Yang, and H.~T. Shen.
\newblock Visual spatial attention network for relationship detection.
\newblock In {\em 2018 ACM Multimedia Conference on Multimedia Conference},
  pages 510--518. ACM, 2018.

\bibitem{hsieh2015pu}
C.-J. Hsieh, N.~Natarajan, and I.~S. Dhillon.
\newblock Pu learning for matrix completion.
\newblock In {\em ICML}, pages 2445--2453, 2015.

\bibitem{hu2018relation}
H.~Hu, J.~Gu, Z.~Zhang, J.~Dai, and Y.~Wei.
\newblock Relation networks for object detection.
\newblock In {\em The IEEE Conference on Computer Vision and Pattern
  Recognition (CVPR)}, June 2018.

\bibitem{hwang2018tensorize}
S.~Jae~Hwang, S.~N. Ravi, Z.~Tao, H.~J. Kim, M.~D. Collins, and V.~Singh.
\newblock Tensorize, factorize and regularize: Robust visual relationship
  learning.
\newblock In {\em The IEEE Conference on Computer Vision and Pattern
  Recognition (CVPR)}, June 2018.

\bibitem{kaji2018multi}
H.~Kaji, H.~Yamaguchi, and M.~Sugiyama.
\newblock Multi task learning with positive and unlabeled data and its
  application to mental state prediction.
\newblock In {\em 2018 IEEE International Conference on Acoustics, Speech and
  Signal Processing (ICASSP)}, pages 2301--2305. IEEE, 2018.

\bibitem{kanehira2016multi}
A.~Kanehira and T.~Harada.
\newblock Multi-label ranking from positive and unlabeled data.
\newblock In {\em Proceedings of the IEEE Conference on Computer Vision and
  Pattern Recognition}, pages 5138--5146, 2016.

\bibitem{krishna2017visual}
R.~Krishna, Y.~Zhu, O.~Groth, J.~Johnson, K.~Hata, J.~Kravitz, S.~Chen,
  Y.~Kalantidis, L.-J. Li, D.~A. Shamma, et~al.
\newblock Visual genome: Connecting language and vision using crowdsourced
  dense image annotations.
\newblock {\em International Journal of Computer Vision}, 123(1):32--73, 2017.

\bibitem{li2009positive}
X.-L. Li, P.~S. Yu, B.~Liu, and S.-K. Ng.
\newblock Positive unlabeled learning for data stream classification.
\newblock In {\em Proceedings of the 2009 SIAM International Conference on Data
  Mining}, pages 259--270. SIAM, 2009.

\bibitem{li2017vip-cnn}
Y.~Li, W.~Ouyang, X.~Wang, and X.~Tang.
\newblock Vip-cnn: Visual phrase guided convolutional neural network.
\newblock {\em Computer vision and pattern recognition}, pages 7244--7253,
  2017.

\bibitem{liang2018visual}
K.~Liang, Y.~Guo, H.~Chang, and X.~Chen.
\newblock Visual relationship detection with deep structural ranking.
\newblock In {\em AAAI Conference on Artificial Intelligence}, 2018.

\bibitem{liang2017deep}
X.~Liang, L.~Lee, and E.~P. Xing.
\newblock Deep variation-structured reinforcement learning for visual
  relationship and attribute detection.
\newblock {\em Computer vision and pattern recognition}, pages 4408--4417,
  2017.

\bibitem{lu2016visual}
C.~Lu, R.~Krishna, M.~S. Bernstein, and L.~Feifei.
\newblock Visual relationship detection with language priors.
\newblock {\em European conference on computer vision}, pages 852--869, 2016.

\bibitem{ng2002discriminative}
A.~Y. Ng and M.~I. Jordan.
\newblock On discriminative vs. generative classifiers: A comparison of
  logistic regression and naive bayes.
\newblock In {\em Advances in neural information processing systems}, pages
  841--848, 2002.

\bibitem{pennington2014glove}
J.~Pennington, R.~Socher, and C.~Manning.
\newblock Glove: Global vectors for word representation.
\newblock In {\em Proceedings of the 2014 conference on empirical methods in
  natural language processing (EMNLP)}, pages 1532--1543, 2014.

\bibitem{peyre2017weakly-supervised}
J.~Peyre, I.~Laptev, C.~Schmid, and J.~Sivic.
\newblock Weakly-supervised learning of visual relations.
\newblock {\em international conference on computer vision}, pages 5189--5198,
  2017.

\bibitem{platanios2017estimating}
E.~Platanios, H.~Poon, T.~M. Mitchell, and E.~J. Horvitz.
\newblock Estimating accuracy from unlabeled data: A probabilistic logic
  approach.
\newblock In {\em Advances in Neural Information Processing Systems}, pages
  4361--4370, 2017.

\bibitem{redmon2016you}
J.~Redmon, S.~Divvala, R.~Girshick, and A.~Farhadi.
\newblock You only look once: Unified, real-time object detection.
\newblock In {\em Proceedings of the IEEE conference on computer vision and
  pattern recognition}, pages 779--788, 2016.

\bibitem{ren2017faster}
S.~Ren, K.~He, R.~Girshick, and J.~Sun.
\newblock Faster r-cnn: towards real-time object detection with region proposal
  networks.
\newblock {\em IEEE Transactions on Pattern Analysis \& Machine Intelligence},
  (6):1137--1149, 2017.

\bibitem{sansone2018efficient}
E.~Sansone, F.~G. De~Natale, and Z.-H. Zhou.
\newblock Efficient training for positive unlabeled learning.
\newblock {\em IEEE Transactions on Pattern Analysis and Machine Intelligence},
  2018.

\bibitem{xu2017scene}
D.~Xu, Y.~Zhu, C.~B. Choy, and L.~Fei-Fei.
\newblock Scene graph generation by iterative message passing.
\newblock In {\em Proceedings of the IEEE Conference on Computer Vision and
  Pattern Recognition}, volume~2, 2017.

\bibitem{yang2018shuffle}
X.~Yang, H.~Zhang, and J.~Cai.
\newblock Shuffle-then-assemble: learning object-agnostic visual relationship
  features.
\newblock {\em arXiv preprint arXiv:1808.00171}, 2018.

\bibitem{yin2018zoom}
G.~Yin, L.~Sheng, B.~Liu, N.~Yu, X.~Wang, J.~Shao, and C.~C. Loy.
\newblock Zoom-net: Mining deep feature interactions for visual relationship
  recognition.
\newblock {\em arXiv preprint arXiv:1807.04979}, 2018.

\bibitem{yu2017visual}
R.~Yu, A.~Li, V.~I. Morariu, and L.~S. Davis.
\newblock Visual relationship detection with internal and external linguistic
  knowledge distillation.
\newblock {\em International conference on computer vision}, pages 1068--1076,
  2017.

\bibitem{zellers2018neural}
R.~Zellers, M.~Yatskar, S.~Thomson, and Y.~Choi.
\newblock Neural motifs: Scene graph parsing with global context.
\newblock In {\em Proceedings of the IEEE Conference on Computer Vision and
  Pattern Recognition}, pages 5831--5840, 2018.

\bibitem{zhan2018comprehensive}
Y.~Zhan, J.~Yu, Z.~Yu, R.~Zhang, D.~Tao, and Q.~Tian.
\newblock Comprehensive distance-preserving autoencoders for cross-modal
  retrieval.
\newblock In {\em 2018 ACM Multimedia Conference on Multimedia Conference},
  pages 1137--1145. ACM, 2018.

\bibitem{zhang2017visual}
H.~Zhang, Z.~Kyaw, S.~Chang, and T.~Chua.
\newblock Visual translation embedding network for visual relation detection.
\newblock {\em Computer vision and pattern recognition}, pages 3107--3115,
  2017.

\bibitem{zhang2017ppr-fcn}
H.~Zhang, Z.~Kyaw, J.~Yu, and S.~Chang.
\newblock Ppr-fcn: Weakly supervised visual relation detection via parallel
  pairwise r-fcn.
\newblock {\em International conference on computer vision}, pages 4243--4251,
  2017.

\bibitem{zhang2017universum}
X.~Zhang and Y.~LeCun.
\newblock Universum prescription: Regularization using unlabeled data.
\newblock In {\em AAAI}, pages 2907--2913, 2017.

\bibitem{zhou2012multi}
J.~T. Zhou, S.~J. Pan, Q.~Mao, and I.~W. Tsang.
\newblock Multi-view positive and unlabeled learning.
\newblock In {\em Asian Conference on Machine Learning}, pages 555--570, 2012.

\bibitem{zhu2018deep}
Y.~Zhu and S.~Jiang.
\newblock Deep structured learning for visual relationship detection.
\newblock In {\em AAAI Conference on Artificial Intelligence}, 2018.

\end{thebibliography}
}

\end{document}